\documentclass[twocolumn]{article}
\usepackage{preprint}
\usepackage{geometry}  
\usepackage{hyperref}
\hypersetup{colorlinks=true, linkcolor=purple, urlcolor=blue, citecolor=cyan, anchorcolor=black}
\usepackage[numbers,square]{natbib}
\usepackage[utf8]{inputenc}
\usepackage[T1]{fontenc}
\usepackage{tikz}
\usepackage{array}
\usepackage{lineno}
\usepackage{xcolor}
\usepackage{amssymb}
\usepackage{authblk}
\usepackage{caption}
\usepackage{amsmath}
\usepackage{booktabs}
\usepackage{tabularx}
\usepackage{float}
\usepackage{titlesec}
\titlespacing\section{0pt}{12pt plus 3pt minus 3pt}{1pt plus 1pt minus 1pt}
\titlespacing\subsection{0pt}{10pt plus 3pt minus 3pt}{1pt plus 1pt minus 1pt}
\titlespacing\subsubsection{0pt}{8pt plus 3pt minus 3pt}{1pt plus 1pt minus 1pt}
\title{Learned Image Compression and Restoration for Digital Pathology}

\author[1]{SeonYeong Lee\textsuperscript{\dag}}
\author[1]{EonSeung Seong\textsuperscript{\dag}}
\author[1]{DongEon Lee}
\author[1]{SiYeoul Lee}
\author[1]{Yubin Cho}
\author[1]{Chunsu Park}
\author[1]{Seonho Kim}
\author[1]{MinKyung Seo}
\author[3]{YoungSin Ko}
\author[1,2,*]{MinWoo Kim}
\affil[1]{Department of Information Convergence Engineering, Pusan National University, Yangsan, Korea}
\affil[2]{School of Biomedical Convergence Engineering, Pusan National University, Yangsan, Korea}
\affil[3]{Seegene Medical Foundation, Seoul, Korea}
\affil[ ]{\textsuperscript{\dag}The first two authors contributed equally to this work.}
\affil[ ]{\textsuperscript{*}Corresponding author: \texttt{mkim180@pusan.ac.kr}}

\begin{document}

\twocolumn[\begin{@twocolumnfalse}
\maketitle
\vspace{-1.0em}
\begin{abstract}
Digital pathology images play a crucial role in medical diagnostics, but their ultra-high resolution and large file sizes pose significant challenges for storage, transmission, and real-time visualization. To address these issues, we propose CLERIC, a novel deep learning-based image compression framework designed specifically for whole slide images (WSIs). CLERIC integrates a learnable lifting scheme and advanced convolutional techniques to enhance compression efficiency while preserving critical pathological details.
Our framework employs a lifting-scheme transform in the analysis stage to decompose images into low- and high-frequency components, enabling more structured latent representations. These components are processed through parallel encoders incorporating Deformable Residual Blocks (DRB) and Recurrent Residual Blocks (R2B) to improve feature extraction and spatial adaptability. The synthesis stage applies an inverse lifting transform for effective image reconstruction, ensuring high-fidelity restoration of fine-grained tissue structures.
We evaluate CLERIC on a digital pathology image dataset and compare its performance against state-of-the-art learned image compression (LIC) models. Experimental results demonstrate that CLERIC achieves superior rate-distortion (RD) performance, significantly reducing storage requirements while maintaining high diagnostic image quality.  
Our study highlights the potential of deep learning-based compression in digital pathology, facilitating efficient data management and long-term storage while ensuring seamless integration into clinical workflows and AI-assisted diagnostic systems.
Code and models are available at: \underline{\url{https://github.com/pnu-amilab/CLERIC}}.
\end{abstract}


\vspace{0.5cm}

\end{@twocolumnfalse}]

\section{Introduction}
\label{sec:Introduction}

Digital pathology images serve as fundamental data for various medical applications, playing a crucial role in cancer diagnosis, disease analysis, and treatment planning. These images are typically stored as Whole Slide Images (WSIs), which are characterized by ultra-high resolution (typically \(\approx 0.25 \mu m/\text{px}\)). A single uncompressed WSI can often exceed several gigabytes in size (e.g., 20–30 GB per image), posing significant challenges in terms of storage, transmission, and computational efficiency. Consequently, there is a growing need for highly efficient image compression techniques to address these issues while preserving critical pathological details.

Most commercial digital scanners rely on standard lossy image compression codecs such as JPEG~\cite{pennebaker1992jpeg} and JPEG2000~\cite{christopoulos2000jpeg2000}. However, these methods require manually tuned parameters, limiting their ability to efficiently handle the complex and intricate structures found in pathology images, particularly at high compression rates. Pathology images contain microscopic, cell-level details and critical structural patterns, and excessive information loss during compression can directly impact diagnostic accuracy. Therefore, an effective compression method must maximize compression efficiency while minimizing distortion in diagnostically relevant regions.

Recently, Learned Image Compression (LIC) methods~\cite{balle2016end,balle2018variational,minnen2018joint,cheng2020learned,minnen2020channel,he2021checkerboard,zou2022devil,he2022elic,jiang2023mlic,jiang2023mlicpp} have outperformed traditional codecs in rate-distortion (RD) performance, establishing them as a promising alternative for medical image storage and transmission. These methods are typically based on Autoencoders (AEs) or Variational Autoencoders (VAEs)~\cite{kingma2013auto} and involve key components such as transform encoding, quantization, entropy coding, and inverse transform decoding. Early studies leveraged CNN-based nonlinear analysis and synthesis transforms to build LIC frameworks. Subsequent advances, such as hyperprior models~\cite{balle2018variational} and context modeling~\cite{minnen2018joint,minnen2020channel,he2021checkerboard}, have significantly improved compression efficiency by enhancing the probability estimation of latent representations and capturing spatial correlations within latent spaces. These techniques have achieved state-of-the-art (SOTA) performance on standard natural image datasets.

Given these advancements, applying LIC to digital pathology images is an attractive research direction. However, relatively few studies have systematically evaluated recent SOTA LIC models on pathology images or tailored their architectures to the unique properties of such data. In this study, we investigate the performance of existing SOTA LIC models on pathological images and introduce a novel compression framework, Content-adaptive Lifting wavelet with Enhanced Residual blocks for Image Compression (CLERIC). Our model builds upon SOTA LIC techniques while incorporating key modifications tailored to pathology images. Specifically, CLERIC enhances the encoder and decoder to better capture spatially complex patterns and tiny structures spanning only a few pixels, which correspond to high-frequency components crucial for pathology. By doing so, CLERIC generates more compact and sparse latent representations, improving quantization and entropy coding efficiency.

To achieve this, our model employs a learnable lifting scheme based on the wavelet transform, which decomposes an image into different frequency components before compression. This separation allows the encoder to effectively extract multiscale features and learn structured latent representations at each scale. The wavelet-based decomposition enhances sparsity, making the representations more compressible. In addition, our model integrates deformable residual blocks and recurrent residual blocks to enhance feature extraction. Standard convolutional layers, commonly used in LIC, suffer from fixed receptive fields, limiting their ability to capture long-range dependencies effectively. To address this, we incorporate deformable convolutions and recurrent convolutions. Deformable convolutions~\cite{dai2017deformable,zhu2019deformable} enhance spatial flexibility by dynamically adjusting sampling locations, allowing the network to better capture complex structures and irregular patterns in pathology images. Meanwhile, recurrent convolutions~\cite{liang2015recurrent} iteratively refine feature representations by reusing convolutional filters across multiple passes, enhancing feature extraction without increasing model complexity. These techniques, originally successful in image segmentation~\cite{alom2018recurrent,shibuya2020feedback} and super-resolution~\cite{kim2016deeply,han2018image,yang2018drfn}, further support their applicability in high-fidelity pathology image compression.

Through these enhancements, our model effectively preserves fine-grained pathological details while achieving superior compression efficiency. We evaluate the rate-distortion performance of CLERIC and compare it with existing state-of-the-art LIC models on pathological image datasets, demonstrating its ability to maintain diagnostic integrity while significantly reducing storage and transmission costs. By integrating wavelet-based decomposition with adaptive and recurrent feature extraction techniques, CLERIC represents a significant advancement in deep learning-based image compression, making it highly applicable to complex medical imaging tasks.

\noindent\textbf{Our main contributions are as follows:}
\setlength{\leftmargini}{15pt}
\begin{itemize}
    \item We propose a novel deep learning-based image compression framework for digital pathology, incorporating a learnable lifting scheme and optimized convolutional blocks to improve feature extraction and latent space compactness.
    \item We systematically evaluate the proposed framework on digital pathology image datasets and demonstrate its superior rate-distortion performance compared to existing state-of-the-art methods.
    \item We show that the proposed compression model supports multi-resolution (pyramidal) formats, making it compatible with standard pathology visualization software and enabling seamless integration into real-world diagnostic workflows.
\vspace{-10pt}
\end{itemize}

\section{Related Works}
\label{sec:Related_Works}

\subsection{Learned Image Compression}
\label{subsec:LIC}

The basic structure of Learned Image Compression (LIC) can be represented as:
\begin{align}
        \label{eq:LIC}
        \bold{y} = g_e(\bold{x}; \boldsymbol{\theta}_{g_e}), \quad \hat{\bold{y}} = Q(\bold{y}), \quad \hat{\bold{x}} = g_d(\hat{\bold{y}}; \boldsymbol{\theta}_{g_d}),
\end{align}
where $\bold{x}$ is the input image, $g_e(\cdot)$ is the encoder that transforms the image space into a latent space with a learnable parameter set $\boldsymbol{\theta}_{g_e}$, and $\bold{y}$ is the latent representation. The quantized representation, $\hat{\bold{y}}$, is obtained through the quantization function $Q(\cdot)$ and is used for entropy coding. The function $g_d(\cdot)$ represents the decoder, which reconstructs the image from the latent representation using a learnable parameter set $\boldsymbol{\theta}_{g_d}$, producing the decompressed image $\hat{\bold{x}}$.

Quantization is inherently non-differentiable, making gradient-based training challenging. To address this, Ballé et al.~\cite{balle2016end} proposed an approach where quantization is approximated by adding uniform noise \( U(-0.5, 0.5) \) during training. Entropy coding, a lossless process, is represented as:
\begin{align}
        \label{eq:entropyC}
        \bold{s} = \eta_e(\hat{\bold{y}}), \quad \hat{\bold{y}} = \eta_d(\bold{s}),
\end{align}
where $\eta_e(\cdot)$ and $\eta_d(\cdot)$ denote the entropy (arithmetic) encoder and decoder, respectively, and $\bold{s}$ is the final compressed bitstream for storage and transmission. Ballé et al.~\cite{balle2016end} also introduced a non-parametric factorized density model to estimate the probability distribution $p_{\hat{\bold{y}}}(\hat{\bold{y}})$, which is crucial for entropy coding. The estimated bit rate is computed as:
\begin{align}
        R(\hat{\bold{y}}) = \mathbb{E}[-\log_2(p_{\hat{\bold{y}}}(\hat{\bold{y}}))].
\end{align}

To further improve compression efficiency, Ballé et al.~\cite{balle2016end} introduced a hyperprior model that captures spatial dependencies within the latent representation $\hat{\bold{y}}$ by incorporating an auxiliary latent variable $\bold{z}$. This conditional entropy model is formulated as:
\begin{align}
        \label{eq:hyperP}
        \bold{z} = h_e(\bold{y}; \boldsymbol{\theta}_{h_e}), \quad \hat{\bold{z}} = Q(\bold{z}), \quad p_{\hat{\bold{y}}|\hat{\bold{z}}}(\hat{\bold{y}}|\hat{\bold{z}} ) \sim h_d(\hat{\bold{z}}; \boldsymbol{\theta}_{h_d}),
\end{align}
where $h_e(\cdot)$ and $h_d(\cdot)$ are the auxiliary encoder and decoder with learnable parameters $\boldsymbol{\theta}_{h_e}$ and $\boldsymbol{\theta}_{h_d}$, respectively. The quantized prior, $\hat{\bold{z}}$, enables a more accurate entropy model, leading to improved compression performance.

In this formulation, each element of $\hat{\bold{y}}$ is modeled as a zero-mean Gaussian, with its probability distribution expressed as:
\begin{align}
        \label{eq:condiEnt}
        p_{\hat{\bold{y}}|\hat{\bold{z}}}(\hat{\bold{y}}|\hat{\bold{z}}) = \prod_i \left(N(\mu_i,\sigma_i^2) * U(-0.5, 0.5)\right)(\hat{y}_i),
\end{align}
where $\mu_i=0$ denotes the zero-mean assumption, and $\sigma_i$ represents the standard deviation (scale) of each latent element, estimated by $h_d(\hat{\bold{z}}; \boldsymbol{\theta}_{h_d})$ as defined in Eq.~\ref{eq:hyperP}. Since $\hat{\bold{z}}$ does not necessarily follow a specific parametric distribution, it is compressed separately using a non-parametric factorized model, similar to $\hat{\bold{y}}$ in models without a hyperprior. Ballé et al. demonstrated that leveraging the additional prior $P(\hat{\bold{z}})$ improves data compression by enhancing entropy modeling, albeit at the cost of additional storage for the hyperprior.

Subsequent research has introduced more accurate entropy models by refining the Gaussian assumption. Minnen et al.~\cite{minnen2018joint} and Lee et al.~\cite{lee2018context} extended this approach by modeling Gaussian distributions with both learnable mean and variance. Further improvements were made by incorporating Gaussian Mixture Models (GMMs)~\cite{cheng2020learned}, providing a more expressive probability distribution.

The entire LIC model is trained using rate-distortion optimization, which minimizes both the expected bit rate and the expected distortion between the original and decompressed images. The loss function for rate-distortion optimization is given by:
\begin{align}
        \label{eq:lossF}
        L = R(\hat{\bold{y}}) + R(\hat{\bold{z}}) + \lambda D(\bold{x}, \hat{\bold{x}}),
\end{align}
where $R(\hat{\bold{y}}) = \mathbb{E}[-\log_2(p_{\hat{\bold{y}}|\hat{\bold{z}}}(\hat{\bold{y}}|\hat{\bold{z}}))]$ and $R(\hat{\bold{z}}) = \mathbb{E}[-\log_2(p_{\hat{\bold{z}}}(\hat{\bold{z}}))]$ denote the bit rates of the latent representation and the hyperprior, respectively. The distortion term, $D(\bold{x}, \hat{\bold{x}}) = \mathbb{E}[\|\bold{x} - \hat{\bold{x}}\|^2]$, represents the mean squared error (MSE) between the original and reconstructed images. The parameter $\lambda$ controls the trade-off between compression efficiency and reconstruction fidelity.

Recent LIC models further enhance entropy coding efficiency by incorporating context models to improve the conditional probability estimation of latent representations~\cite{minnen2018joint}. Context models exploit correlations across different dimensions, including local spatial context, global spatial context, and channel-wise dependencies, to refine probability predictions. The estimation of Gaussian parameters for entropy coding of the latent representation $\hat{\bold{y}}=\{\hat{y}_i^k\}$ can be expressed as:
\begin{align}
        \label{eq:contextM}
        (\boldsymbol{\mu}, \boldsymbol{\sigma}) = \{\mu_i^k, \sigma_i^k\} =  f(\boldsymbol{\psi}, \boldsymbol{\phi}_{(1)}, \boldsymbol{\phi}_{(2)}, \dots ;\boldsymbol{\theta}_{f}),  \\ \nonumber
\boldsymbol{\psi} = h_d(\hat{\bold{z}}; \boldsymbol{\theta}_{h_d}), \quad \boldsymbol{\phi}_{(j)} = g_{(j)}(\hat{\bold{y}}_{(j)}; \boldsymbol{\theta}_{g_{(j)}}),
\end{align}
where $i$ and $k$ denote spatial and channel indices, respectively. The term $\boldsymbol{\psi}$ represents the hyperprior representation, while $g_{(j)}(\cdot)$ denotes the $j$th function (module) used to extract the context $\boldsymbol{\phi}_{(j)}$ from a slice, part, or chunk of $\hat{\bold{y}}$, with learnable parameters $\boldsymbol{\theta}_{g_{(j)}}$. The function $f(\cdot)$ then combines all available information to output the final Gaussian parameters, with learnable parameters $\boldsymbol{\theta}_{f}$.
 This follows the information theory principle:
\begin{align}
\label{eq:infoTh}
\mathbb{E}[-\log_2(p_{\hat{\bold{y}}|\hat{\bold{z}}}(\hat{\bold{y}}|\hat{\bold{z}}))]
=\mathbb{E}[-\log_2(p_{\hat{\bold{y}}|\boldsymbol{\psi}}(\hat{\bold{y}}|\boldsymbol{\psi}))]  \\ \nonumber
\leq \mathbb{E}[-\log_2(p_{\hat{\bold{y}}|\{\boldsymbol{\psi}, \boldsymbol{\phi}_{(1)}, \boldsymbol{\phi}_{(2)}, \dots\}}(\hat{\bold{y}}|\{\boldsymbol{\psi}, \boldsymbol{\phi}_{(1)}, \boldsymbol{\phi}_{(2)}, \dots\}))].
\end{align}

Early approaches employed autoregressive (AR) context models, which estimate probability distributions sequentially based on the spatial neighborhood of each latent symbol, thereby maximizing entropy coding efficiency. However, AR-based models inherently suffer from sequential processing constraints, limiting their practical speed. To address this limitation, recent models process latent representations in parallel by dividing them into multiple independent groups. The ChARM (Channel-wise Autoregressive Model) introduced a strategy to split latents along the channel dimension to achieve parallelism while retaining predictive efficiency~\cite{minnen2020channel}. Similarly, the Checkerboard Context Model spatially partitions latents into non-overlapping subgroups, enabling efficient parallel decoding~\cite{he2021checkerboard}. A more advanced approach, SCCTX (Space-Channel Context Model), integrates both spatial and channel-wise context modeling, further improving compression efficiency~\cite{he2022elic}. Recent studies have extended these approaches by capturing inter-slice dependencies within multi-slice latent representations, leading to even more efficient entropy models~\cite{jiang2023mlic,jiang2023mlicpp}. These methods consider not only intra-slice spatial and channel correlations but also cross-slice correlations, enhancing overall rate-distortion performance.

\subsection{Pathological Image Compression}
\label{subsec:LIC_Path}

Despite significant advancements in image compression technology, most whole slide imaging (WSI) vendors continue to rely on the JPEG algorithm, which was originally developed nearly five decades ago. Beyond the conventional JPEG compression, some vendors have adopted JPEG2000, which replaces the Discrete Cosine Transform (DCT) with the Discrete Wavelet Transform (DWT). This shift enables more efficient lossy compression while maintaining better flexibility in scalable bit-rate control. However, even with these improvements, model-based compression techniques such as JPEG2000 remain fundamentally limited in adapting to the intricate structures and high-dimensional feature distributions present in pathological images.

In natural images, deep learning-based learned image compression (LIC) has demonstrated significantly better compression efficiency than traditional model-based methods like JPEG2000. However, despite these advancements, only a limited number of studies have explored deep learning-based methods for pathological image compression. Among them, Fischer et al. recently introduced Stain Quantized Latent Compression (SQLC)\cite{fischer2024learned}, a deep learning-based approach that first encodes staining and RGB channels (6 channels) into 3 latent channels before applying a compression autoencoder. While this method outperforms traditional approaches such as JPEG in classification tasks and largely preserves image quality metrics, the effect of the stain encoding process alone has not been rigorously studied, and the compression model itself is based on an early LIC framework without hyperprior modeling.

Despite the clear potential of deep learning in pathological image compression, no study has comprehensively investigated or adapted current state-of-the-art (SOTA) LIC models for histopathological images. The unique challenges posed by high-resolution, multi-scale structures, and clinically significant small details in pathology images necessitate a more tailored adaptation of modern LIC frameworks. This gap underscores the need for further research to modify and optimize SOTA LIC architectures to effectively address the specific demands of pathological imaging.

\section{Methods}
\label{sec:Methods}

\begin{figure*}[t]
        \captionsetup{labelfont=bf}
        \centerline{\includegraphics[width=\textwidth ,height= 7cm]{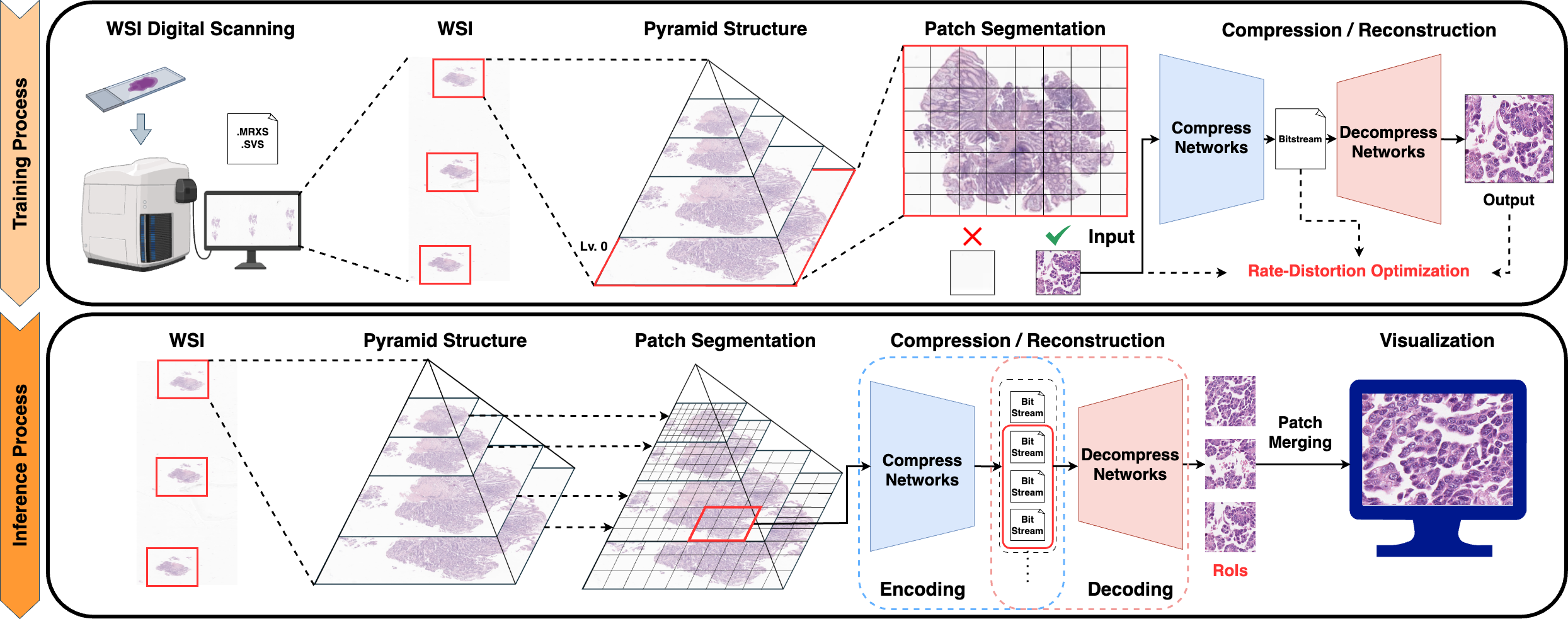}}
        \caption{Overview of the proposed workflow for pathology image compression. Whole slide images (WSIs) are digitized, segmented into patches, and processed through compression and decompression models. The framework supports multi-resolution storage and retrieval, ensuring efficient image reconstruction and real-time visualization in pathology software.}
        \label{fig:workflow}
        \vspace{-15pt}
\end{figure*}

\subsection{LIC for Pathological Images}
\label{subsec:LICPath}

Figure~\ref{fig:workflow} illustrates the overall workflow of our proposed method. We collected whole slide images (WSIs) digitized by commercial scanners. Each scanner selectively scans local regions containing one or more dyed tissue samples, and these regions are stored in multi-resolution formats (pyramid structures) such as \texttt{.mrxs} and \texttt{.svs}. In this hierarchical format, Level 0 represents the full-resolution image, while Level $n$ denotes an image downsampled from Level $n-1$. 

For model training, we extracted full-resolution (Level 0) images using OpenSlide, an open-source library for handling WSIs. Each WSI was segmented into non-overlapping patches of size $H \times W$. During preprocessing, we discarded patches containing only the white background (empty regions) or those without any tissue content. The remaining patches were used to train the image compression (encoding) and decompression (decoding) models. 
In the encoding process, each patch was passed through the compression model, which output a corresponding bitstream file. Conversely, in the decoding process, the decompression model took a bitstream as input and reconstructed the corresponding decompressed image.

For the inference phase, we used independent WSI datasets that were not included in the training process. Similar to training, we extracted full-resolution (Level 0) images, segmented them into patches, and selected only patches containing tissue regions. Each selected patch was then processed through the trained models to evaluate compression and reconstruction performance.
Additionally, we extended the testing process to include downsampled WSI images from lower resolutions (Levels $1, 2, 3, \dots$). Although the models were trained exclusively on full-resolution patches, this experiment was conducted to assess the model’s generalization capability for lower-resolution image compression and restoration. This evaluation demonstrates the practicality of our approach in real-world software applications.

In a real-world pathology workflow, software tools must efficiently manage WSIs across multiple resolution levels. Ideally, the compression model should compress patches from all levels, enabling an optimized storage and retrieval system. When a pathologist interacts with the WSI viewer, zooming in and out to examine specific regions, the software must efficiently retrieve and decompress only the necessary patches corresponding to the selected resolution and region of interest. The decompression model reconstructs the required patches in real-time for visualization on the monitor, ensuring seamless navigation across different magnification levels. This capability highlights the practical applicability of our approach in digital pathology software by enabling efficient storage and dynamic visualization.

\begin{figure*}[t]
        \captionsetup{labelfont=bf}
        \centerline{\includegraphics[width=\textwidth]{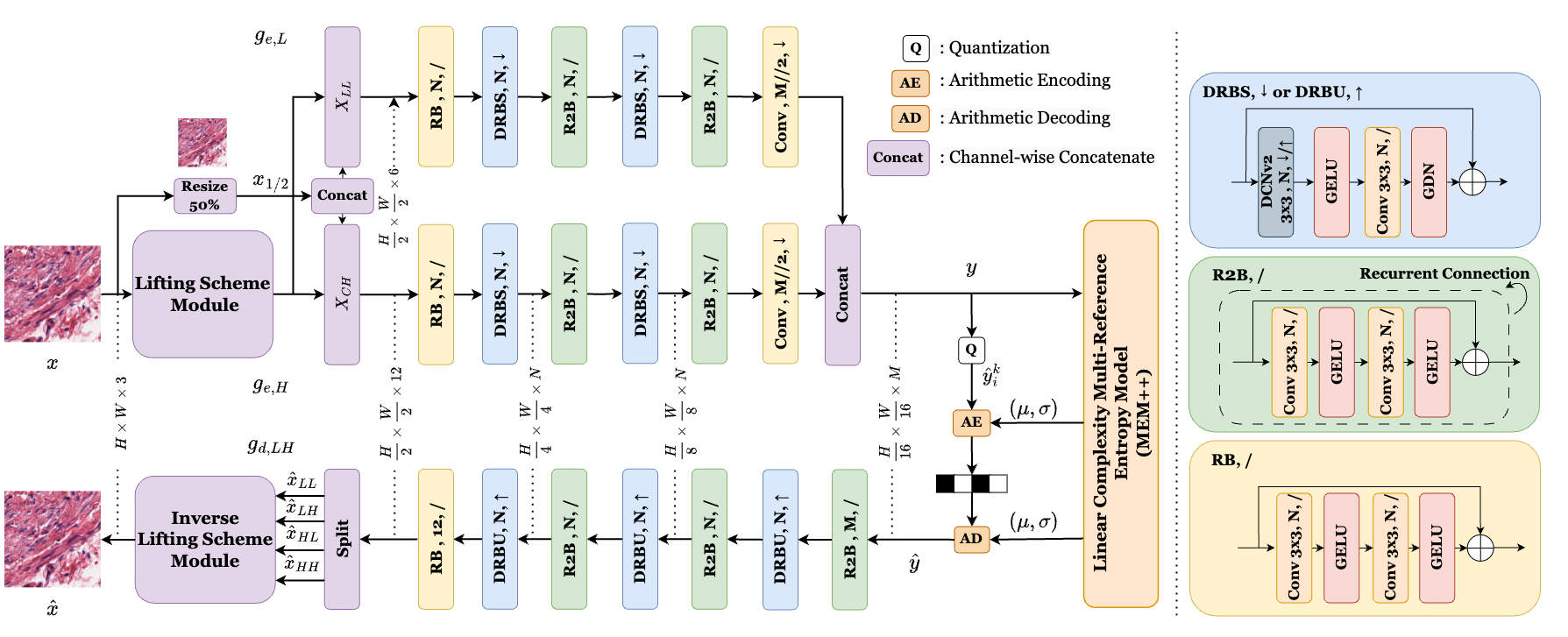}}
        \caption{Overview of the CLERIC framework. The model utilizes a learnable lifting scheme to decompose input images into frequency components, followed by separate encoding branches for low- and high-frequency features. Both the encoder and decoder incorporate deformable and recurrent residual blocks to enhance feature extraction and reconstruction. The compressed latent representation undergoes entropy coding, and the decoder reconstructs the image while preserving fine-grained pathological details.}
        \label{fig:CLERIC_framework}
        \vspace{-15pt}
\end{figure*}

\subsection{CLERIC Model}
\label{subsec:CLERIC_Model}

Figure~\ref{fig:CLERIC_framework} illustrates our CLERIC model, which builds upon MLIC++, the current state-of-the-art (SOTA) model in natural image compression. This choice was made after training and evaluating various existing LIC models on pathological images, where MLIC++ consistently achieved the best rate-distortion (RD) performance.

The encoder $g_e(\bold{x})$ begins with a learnable lifting scheme (wavelet) module, which decomposes each input patch $\bold{x} \in \mathbb{R}^{H \times W \times 3}$ into a low-frequency component $\bold{x}_{LL}   \in \mathbb{R}^{H/2 \times W/2 \times 3}$ and a concatenation of high-frequency components $\bold{x}_{CH} = [\bold{x}_{LH}, \bold{x}_{HL}, \bold{x}_{HH}] \in \mathbb{R}^{H/2 \times W/2 \times 9}$. The subsequent part of the encoder consists of two separate branches, $g_{e,L}(\cdot)$ and $g_{e,H}(\cdot)$, designed to process the low- and high-frequency components independently. This explicit frequency decomposition improves feature extraction efficiency by allowing independent processing of different spectral components.

To integrate spatial and frequency information, we combine the low- and high-frequency components with a downsampled version of the input patch, denoted as $\bold{x}_{1/2}$. This results in the augmented representations:
\begin{equation}
        \begin{aligned}
            \tilde{\bold{x}}_{L} &= [\bold{x}_{LL}, \bold{x}_{1/2}] 
            \in \mathbb{R}^{H/2 \times W/2 \times 6}, \\
            \tilde{\bold{x}}_{H} &= [\bold{x}_{CH}, \bold{x}_{1/2}] 
            \in \mathbb{R}^{H/2 \times W/2 \times 12}.
        \end{aligned}
    \end{equation}
    
The concatenated representations are then fed into their respective branches, and their outputs are combined as:
\begin{equation}
    \bold{y} = [g_{e,L}(\tilde{\bold{x}}_{L}), g_{e,H}(\tilde{\bold{x}}_{H})] \in \mathbb{R}^{H/16 \times W/16 \times M}.
\end{equation}
Since the lifting scheme module may lead to spatial information loss, including the raw downsampled patch $\bold{x}_{1/2}$ helps retain global spatial structure, complementing the frequency-domain features and improving overall feature representation. Empirically, this approach reduces encoder complexity while preserving performance.

To further enhance feature extraction, each branch of the encoder is equipped with deformable residual blocks and recurrent residual blocks. Deformable residual blocks allow the network to adaptively adjust receptive fields, effectively capturing complex spatial patterns in pathology images. Recurrent residual blocks refine feature representations through iterative feedback, improving feature compactness and sparsity in the latent space. These additions facilitate more efficient entropy coding by producing a more structured and compressible latent representation $\bold{y}$.

For entropy coding, we employ the multi-reference entropy model (MEM++) from MLIC++, which has linear computational complexity while efficiently encoding the latent representation $\bold{y}$ into a compressed bitstream. The entropy decoder reconstructs the latent representation $\hat{\bold{y}} \in \mathbb{R}^{H/16 \times W/16 \times M}$, which is then passed to the decoder $g_d(\hat{\bold{y}})$.

The decoder is designed as the inverse counterpart of the encoder. A natural approach would be to split $\hat{\bold{y}}$ into two branches, mirroring $g_{e,L}(\cdot)$ and $g_{e,H}(\cdot)$. However, empirical results show that performance is maintained even with a single unified decoder $g_{d,LH}(\cdot)$, leading to a more efficient architecture. Thus, we process the latent representation as:
\begin{equation}
    \hat{\bold{x}}_{LH} = g_{d,LH}(\hat{\bold{y}}) \in \mathbb{R}^{H/2 \times W/2 \times 12}.
\end{equation}
The decoder then splits $\hat{\bold{x}}_{LH}$ into its respective components, $\hat{\bold{x}}_{LL}, \hat{\bold{x}}_{LH}, \hat{\bold{x}}_{HL}, \hat{\bold{x}}_{HH} \in \mathbb{R}^{H/2 \times W/2 \times 3}$.
These components are then fed into the inverse lifting scheme module, which reconstructs the final estimate of the original patch $\hat{\bold{x}} \in \mathbb{R}^{H/2 \times W/2 \times 3}$.

Similar to the encoder, the decoder is equipped with deformable residual blocks and recurrent residual blocks to enhance the reconstruction quality. Deformable residual blocks help adaptively reconstruct spatial details, while recurrent residual blocks refine feature reconstruction iteratively, allowing more accurate restoration of fine-grained structures.

All model parameters are optimized by minimizing the rate-distortion loss, ensuring efficient compression while preserving image quality.

\subsection{Learned Lifting Scheme Module}
\label{subsec:liftingScheme}

\begin{figure}
        \captionsetup{labelfont=bf}
        \centerline{\includegraphics[width=\columnwidth]{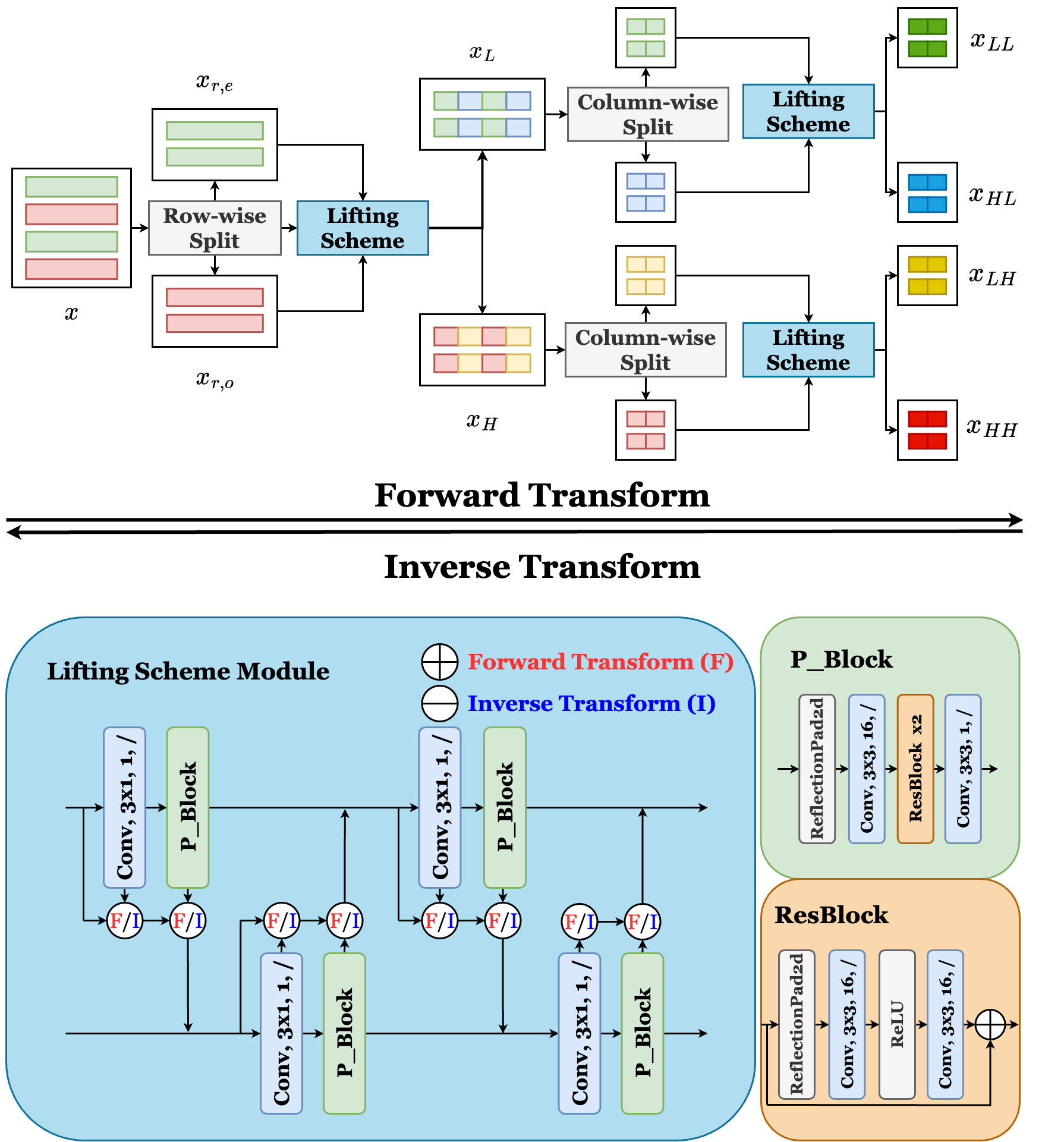}}
\vspace{-0pt}
        \caption{Illustration of the learned lifting scheme module. The input image is progressively decomposed into low- and high-frequency components using a row-wise and column-wise lifting scheme. The resulting four sub-bands capture multiscale structural information, facilitating more efficient compression and reconstruction. The module employs the Cohen-Daubechies-Feauveau (CDF) 9/7 wavelet and learnable convolutional operators to refine the decomposition process.}
\vspace{-10pt}
        \label{fig:liftingscheme_2D}
\end{figure}

Our encoder incorporates a learned lifting scheme, as illustrated in Figure.~\ref{fig:liftingscheme_2D}. Ma et al.~\cite{ma2020end} previously introduced the lifting scheme for natural image compression. Their framework consists solely of forward learned lifting schemes and their perfectly invertible counterparts, enabling both lossless and lossy compression. Lossless compression is supported due to the reversibility of the lifting scheme, while lossy compression is achieved by applying quantization and adjusting the quantization scale. However, despite its versatility, this approach exhibits relatively low rate-distortion (RD) performance compared to state-of-the-art (SOTA) models based on variational autoencoders (VAEs). 

As shown in Figure~\ref{fig:CLERIC_framework}, our lifting scheme serves as an auxiliary module that decomposes the image into multiple frequency bands, thereby assisting the subsequent neural network blocks in producing structured latent representations at different scales. This hierarchical decomposition enables the model to effectively capture both coarse and fine-grained details, improving compression efficiency.

Figure~\ref{fig:liftingscheme_2D} illustrates the lifting scheme-based wavelet module. First, the input image $\bold{x} \in \mathbb{R}^{H \times W \times 3}$ is row-wise split into $\bold{x}_{r,o} \in \mathbb{R}^{H/2 \times W \times 3}$ and $\bold{x}_{r,e} \in \mathbb{R}^{H/2 \times W \times 3}$, where $\bold{x}_{r,o}$ and $\bold{x}_{r,e}$ contain the odd- and even-indexed rows, respectively. This lifting scheme first predicts the odd samples $\bold{x}_{r,o}$ using the even samples $\bold{x}_{r,e}$ through a predict operator $P$. Subsequently, the even samples $\bold{x}_{r,e}$ are updated using the update operator $U$, which operates on the odd samples produced by the predict step. 

In our module, these two steps alternate twice to produce the low-pass and high-pass bands, denoted as $\bold{x}_L$ and $\bold{x}_H$. The subsequent step involves column-wise splitting of $\bold{x}_L$ and $\bold{x}_H$, followed by another lifting scheme application along the column dimension. This process ultimately results in four sub-bands: low-low ($\bold{x}_{LL}$), high-low ($\bold{x}_{HL}$), low-high ($\bold{x}_{LH}$), and high-high ($\bold{x}_{HH}$), capturing information at different frequency scales.

For the lifting scheme, we employ the Cohen-Daubechies-Feauveau (CDF) 9/7 wavelet~\cite{daubechies1998factoring}, which is a core component of the JPEG2000 standard. This is implemented using convolution operations with pre-defined filter coefficients. As shown in Figure~\ref{fig:liftingscheme_2D}, the learnable blocks in our model further refine the wavelet decomposition, making it more suitable for image compression. To enhance structural consistency, we use a shared neural network structure with shared parameters for both the predict and update operators, $P$ and $U$.

\subsection{Deformable Residual Block (DRB)}
\label{subsec:Deformable_Residual_Blocks}

\begin{figure}
        \captionsetup{labelfont=bf}
        \centerline{\includegraphics[width=\columnwidth ,height =5cm]{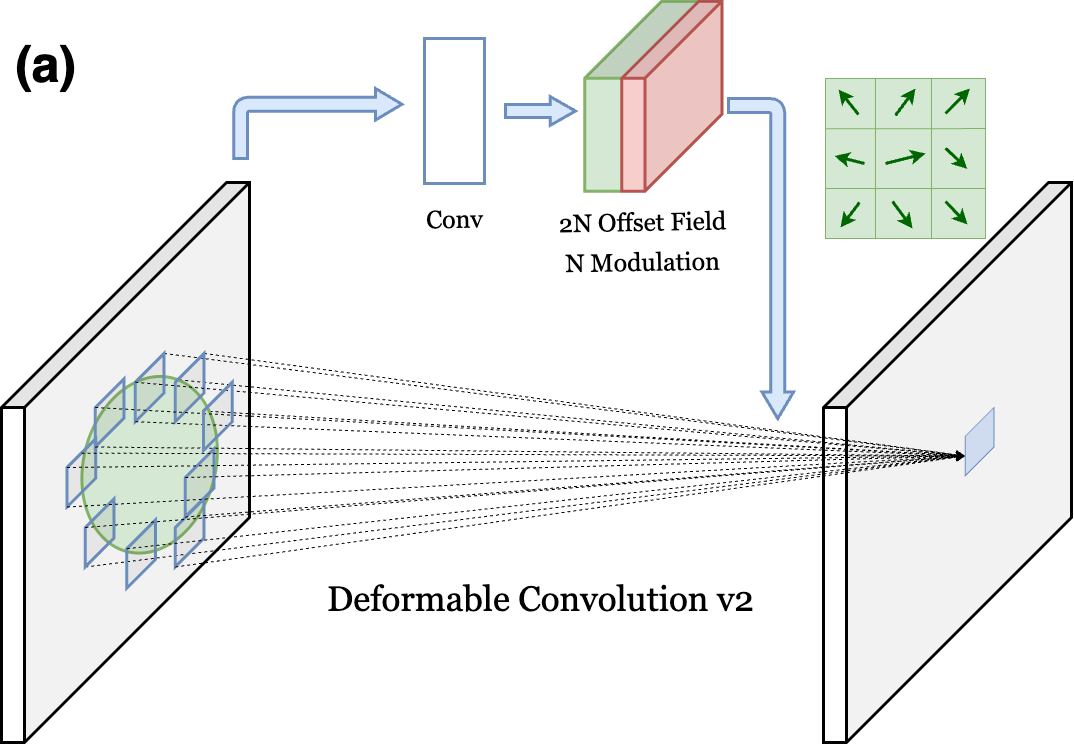}}
        \captionsetup{labelfont=bf}
        \centerline{\includegraphics[width=\columnwidth ,height =4.8cm]{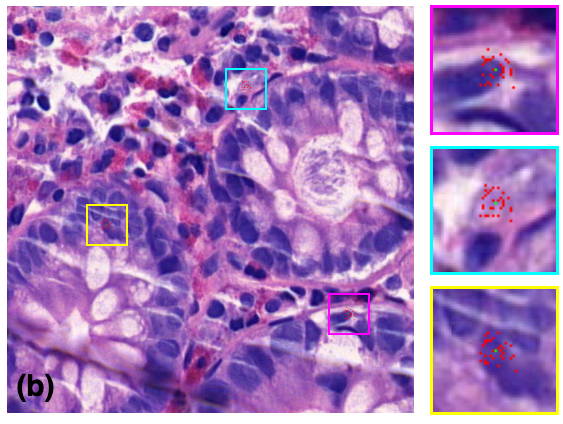}}
\vspace{-5pt}
        \caption{(a) Structure of Deformable Convolution v2 (DCNv2). DCNv2 enhances standard convolution by introducing learnable offsets and modulation scalars, allowing adaptive sampling and improved spatial feature extraction. (b) Visualization of learned offset sampling positions in the Deformable Residual Block with Stride (DRBS). The dynamically adjusted receptive fields help preserve structural integrity and improve feature representation, particularly in pathology image compression.}
\vspace{-15pt}
        \label{fig:DCNv2}
\end{figure}

Deformable Convolutional Networks (DCN) enhance standard convolution by replacing fixed sampling positions with learnable offsets, enabling the network to adaptively capture spatially irregular features. This flexibility allows DCN to outperform conventional convolutions in tasks such as object detection, segmentation, and image classification, particularly in scenarios involving complex structures and variable spatial distributions, while introducing minimal additional computational overhead.

An improved version, Deformable Convolution v2 (DCNv2)~\cite{zhu2019deformable}, extends DCN by incorporating learnable modulation scalars in addition to the learnable offsets. These modulation scalars act as adaptive weights for each sampling position, increasing the representational capacity of the network. As shown in Figure~\ref{fig:DCNv2} (a), the learnable offset field dynamically adjusts sampling positions, while the modulation scalars refine feature weighting to improve spatial information processing. The DCNv2 operation is expressed as:
\begin{equation}
    y(p) = \sum_{k=1}^{K} w_k \cdot x(p + p_k + \Delta p_k) \cdot \Delta m_k,
\end{equation}
where \( \Delta p_k \) and \( \Delta m_k \) are the learnable offset and modulation scalar for the \(k\)-th location, respectively. The modulation scalar \(\Delta m_k\) lies in the range \([0, 1]\), while the offset \(\Delta p_k\) is a real-valued parameter. Fractional offsets are handled via bilinear interpolation to ensure smooth feature extraction.

The ability of DCNv2 to dynamically adjust receptive fields facilitates the generation of sparse and structured latent representations. This characteristic is particularly beneficial in pathology image compression, where certain regions contain critical diagnostic information while others are less relevant. By adaptively prioritizing important spatial regions, DCNv2 enhances feature extraction and compression efficiency.

In our framework, we incorporate DCNv2 into Deformable Residual Blocks (DRBs) to improve both encoding and decoding. Specifically, we introduce two variants: the Deformable Residual Block with Stride (DRBS) for downsampling in the encoder and the Deformable Residual Block with Upsample (DRBU) for upsampling in the decoder. These blocks extend the residual block architecture of the baseline model~\cite{jiang2023mlicpp} and mitigate the limitations of fixed-grid convolutions by enabling dynamic feature sampling.

DRBS effectively reduces spatial resolution while preserving fine-grained details by learning offsets and modulation parameters that guide sampling during downsampling. Figure~\ref{fig:DCNv2} (b) visualizes the learned offset sampling positions in DRBS, demonstrating how DCNv2 dynamically adapts receptive fields to maintain structural integrity. Compared to conventional stride-based downsampling, this approach better retains spatial information critical for pathology image reconstruction.

In the decoder, DRBU is designed to perform upsampling using DCNv2-based Sub-Pixel Convolution instead of conventional convolution-based Sub-Pixel Convolution. This establishes a symmetric structure with the DCNv2-based downsampling blocks in the encoder, while DCNv2 dynamically adjusts sampling positions for more precise and flexible upsampling.

\subsection{Recurrent Residual Blocks (R2B)}
\label{subsec:Recurrent_Residual_Block}

Recurrent Convolution~\cite{liang2015recurrent} enhances representation learning by reusing the same network weights across multiple iterations, enabling feature refinement without increasing the number of parameters. Unlike conventional CNNs that propagate information in a purely feedforward manner, recurrent convolution introduces a feedback mechanism where output features are reintroduced as input while maintaining weight sharing. This approach expands the receptive field and allows for more effective feature representation learning. Due to its efficiency, recurrent convolution has been widely used in various imaging tasks, including segmentation and super-resolution~\cite{alom2018recurrent, shibuya2020feedback, mubashar2022r2u++, kim2016deeply, han2018image, yang2018drfn}. 

To the best of our knowledge, recurrent residual learning has not yet been explored in the context of Learned Image Compression (LIC). In this study, we integrate it into residual blocks to enhance feature refinement and improve compression performance.

To improve the efficiency of the baseline residual block, we design the Recurrent Residual Block (R2B), which incorporates a recursive mechanism for iterative feature refinement. 

R2B extends the conventional residual block by introducing a recurrent structure, where the output features are repeatedly fed back as input. Instead of stacking multiple layers to increase the depth of the network, R2B applies the same convolutional transformation iteratively, enabling progressive feature enhancement while maintaining a compact architecture. 

The operations in R2B are defined as follows:
\begin{equation}
        \begin{aligned}
            X_t &= F(X_{t-1}) + X_{t-1}, \quad t = 1,2, \dots \\ 
            F(X) &= \text{GELU}(\text{Conv}_2(\text{GELU}(\text{Conv}_1(X)))).
        \end{aligned}
    \label{eq:r2b}
\end{equation}

Equation (\ref{eq:r2b}) describes the iterative process of R2B. The first equation defines the recursive update, where the transformed feature \( F(X_{t-1}) \) is added to the previous feature map \( X_{t-1} \), producing an updated representation \( X_t \). This recursive formulation allows for continuous feature refinement over multiple iterations \( t \), enabling more effective learning of hierarchical image structures. By leveraging weight sharing across iterations, R2B maintains parameter efficiency while improving the quality of learned features.

\section{Experimental Results}

\subsection{Training Dataset}
\label{subsec:Data_acq}

Anonymized 48 whole slide images (WSIs) were obtained from Seegen Medical Foundation using two scanners: Pannoramic 250 (3D Histech, Hungary) and Aperio GT450 (Leica Biosystems, Germany). The respective file formats were \texttt{.mrxs} and \texttt{.svs}, with scanning magnifications of $20\times$ and $40\times$, respectively. The dataset includes diverse tissue samples from the cecum, colon, rectum, and stomach.

This study was approved by the Institutional Review Board (IRB) of Pusan National University, and the requirement for informed consent was waived due to the retrospective nature of the study and the use of anonymized clinical data. In addition to the proprietary dataset, we incorporated an open-source dataset~\cite{wang2021dataset}, consisting of 22 H\&E-stained WSIs, to enhance the generalizability of our model.

For model training, we randomly selected 70 WSIs. To mitigate data bias, we uniformly extracted 300 patches of size $256 \times 256$ ($H=256$, $W=256$) from each slide. In total, 21,000 patches were collected, with 20,000 used for training and the remaining 1,000 set aside for validation.

To evaluate rate-distortion (RD) performance, we randomly selected 30 test patches, including 15 patches from \texttt{.svs} WSIs, 15 patches from \texttt{.mrxs} WSIs, and 15 patches from the open-source dataset. Each patch was of size $512 \times 512$ ($H=512$, $W=512$) and was randomly extracted from a different WSI, ensuring that no test patches overlapped with the training or validation datasets.

The patch sizes during training and testing were determined based on computational efficiency, resource availability, and comparability with other models. Empirically, we observed that the patch size used during training does not significantly impact RD performance in a statistically meaningful way.

\subsection{Experimental Settings}
\label{sec:Experiments_Settings}

The proposed framework was implemented using PyTorch~\cite{paszke2019pytorch} and CompressAI~\cite{bégaint2020compressaipytorchlibraryevaluation}. Model training was conducted on NVIDIA RTX A5000 and A100 GPUs with a batch size of 8. The Adam optimizer was employed for optimization, with hyperparameters $\beta_1=0.9$ and $\beta_2=0.999$. 

Training was performed for a total of 300 steps, starting with an initial learning rate of $1\times10^{-4}$. The learning rate was gradually decreased, being set to $1\times10^{-5}$ at 50 steps and $1\times10^{-6}$ at 100 steps. To prevent overfitting, early stopping was applied with a patience of 30 steps, monitoring the validation loss. The model was optimized using rate-distortion (RD) minimization with Mean Squared Error (MSE) as the distortion metric. 

The quality factors ($\lambda$) for the proposed model were set to $\{0.005, 0.0075, 0.011, 0.02, 0.0335\}$. As shown in Figure~\ref{fig:workflow}, the number of channels in the encoder and decoder was set to $N=192$, $M=320$, and the number of recursions in the recurrent residual block (R2B) was set to $t=2$. Other hyperparameters in the entropy model follow the settings in~\cite{jiang2023mlicpp}.

\subsection{Rate-Distortion Performance}
\label{subsec:RD_Performance}

\begin{figure}
        \captionsetup{labelfont=bf}
        \centerline{\includegraphics[width=\columnwidth ,height=5.5cm ]{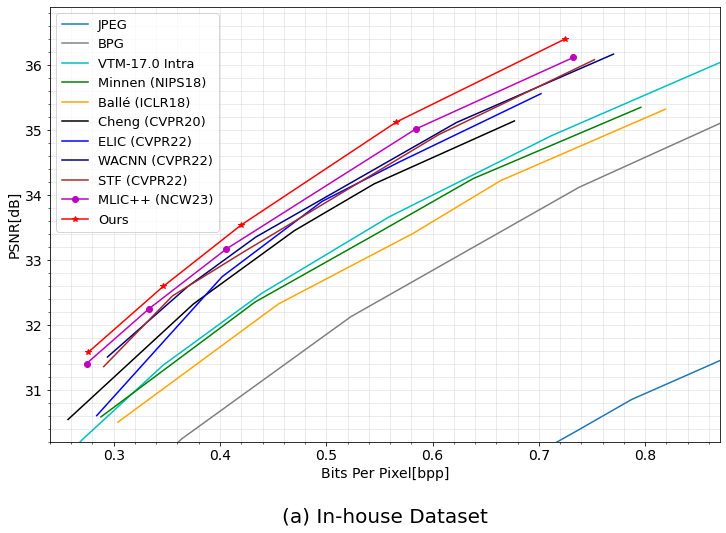}}
        \captionsetup{labelfont=bf}
        \centerline{\includegraphics[width=\columnwidth  ,height=5.5cm ]{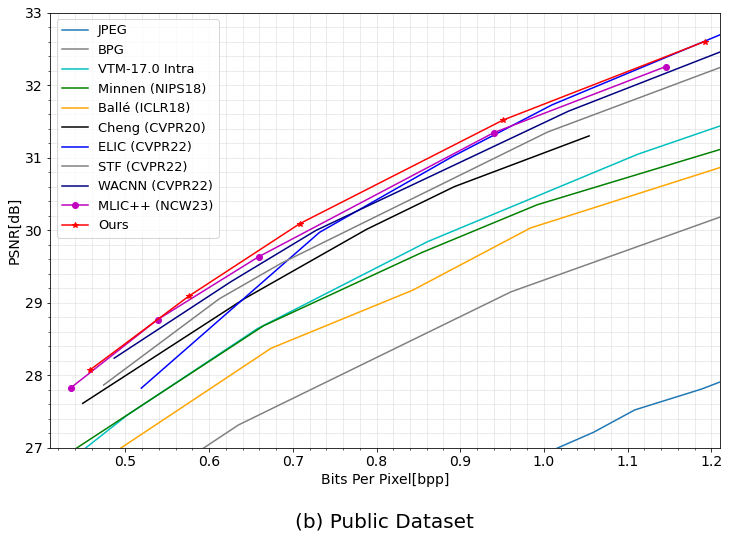}}
\vspace{-5pt}
        \caption{(a) Rate-distortion curves for the proposed CLERIC model and comparison methods on the in-house dataset. (b) Rate-distortion curves for the proposed CLERIC model and comparison methods on the public dataset. Higher curves indicate better compression efficiency and reconstruction quality.}
\vspace{-15pt}
        \label{fig:RD_Curves}
\end{figure}

We compared our model with the current state-of-the-art (SOTA) learned image compression (LIC) methods~\cite{balle2018variational,minnen2018joint,cheng2020learned,he2022elic,zou2022devil,jiang2023mlicpp} as well as conventional compression methods, including JPEG, BPG, and VTM 17.0 Intra. Figure~\ref{fig:RD_Curves} (a)and (b) present the rate-distortion (RD) curves obtained as the sample means of the test patch results for all test patches, \texttt{.svs} test patches, and \texttt{.mrxs} test patches, respectively. 

Our proposed model, CLERIC, outperforms all comparison methods, achieving higher compression efficiency and superior reconstruction quality across all quality factors ($\lambda$). 
Notably, our model achieves a performance improvement of 0.15 dB to 0.25 dB over MLIC++. Since both CLERIC and MLIC++ utilize the same entropy model, MEM++, this improvement is primarily attributed to the effectiveness of our advanced encoder and decoder, which incorporate the Lifting Scheme module, Deformable Residual Block, and Recurrent Residual Block.
\begin{table}
        \centering
        \caption{BD-Rate (\%) comparison of different LIC methods and traditional codecs on an in-house dataset, using VTM 17.0 Intra as anchor, where lower values indicate better compression efficiency.}
        \resizebox{\columnwidth}{!}{
        \label{tab:bd-rate}
        \small
        \renewcommand{\arraystretch}{1.0}
        \setlength{\tabcolsep}{6pt}
        \begin{tabular}{l r | l r}
            \toprule
            \textbf{Methods} & \textbf{BD-Rate (\%)} & \textbf{Methods} & \textbf{BD-Rate (\%)} \\
            \midrule
            VTM 17.0 Intra \cite{bross2021overview} & - & ELIC \cite{he2022elic} & -12.39\% \\
            BPG & 16.50\% & STF \cite{zou2022devil} & -16.08\% \\
            JPEG & 113.84\% & WACNN \cite{zou2022devil} & -16.71\% \\
            Ballé \cite{balle2018variational} & 6.90\% & MLIC++ \cite{jiang2023mlicpp} & -19.78\% \\
            Minnen \cite{minnen2018joint} & 1.20\% & \textbf{Ours} & \textbf{-23.09\%} \\
            Cheng \cite{cheng2020learned} & -11.85\% &  &  \\
            \bottomrule
        \end{tabular}
        }
        \vspace{-10pt}
\end{table}
    
We also evaluate the quantitative performance using the Bj\o ntegaard delta rate (BD-Rate)~\cite{bjontegaard2001calculation} computed from the RD curve. Here, VTM-17.0 Intra (YUV444 mode) is used as the anchor. As shown in Table~\ref{tab:bd-rate}, CLERIC achieves a BD-Rate reduction of approximately -23.09\% compared to VTM-17.0 Intra, -3.31\% compared to MLIC++, and -136.93\% compared to JPEG. These results highlight the efficiency of our proposed model in learned image compression.

\begin{figure}
        \captionsetup{labelfont=bf}
        \centerline{\includegraphics[width=\columnwidth]{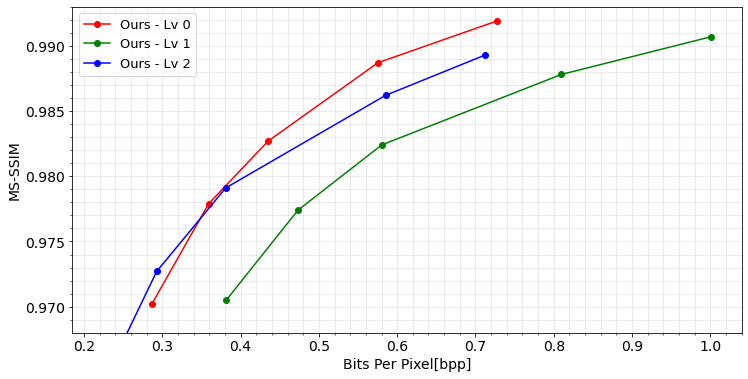}}
\vspace{-0pt}
        \caption{R-D Performance on in-house \texttt{.svs} dataset across pyramid levels.}
\vspace{-15pt}
        \label{fig:multi_scale_ssim}
\end{figure}
Figure~\ref{fig:multi_scale_ssim} shows the rate-distortion performance of our compression model on randomly extracted 512×512 patches from different pyramid levels of an in-house \texttt{.svs} evaluate dataset. Although the model was trained using only level 0(×1) patches, it maintains a similar MS-SSIM scale across levels 1(×4) and 2(×16), demonstrating its ability to preserve structural similarity even at different resolutions. For evaluation, we randomly extracted two patches per slide from 15 pathology slides, resulting in 30 patches per level, which were used to assess the model’s performance across pyramid levels.

\subsection{Qualitative Performance}
\label{subsec:Q_Performance}
\begin{figure}[t]
        \captionsetup{labelfont=bf}
        \centerline{\includegraphics[width=\columnwidth ]{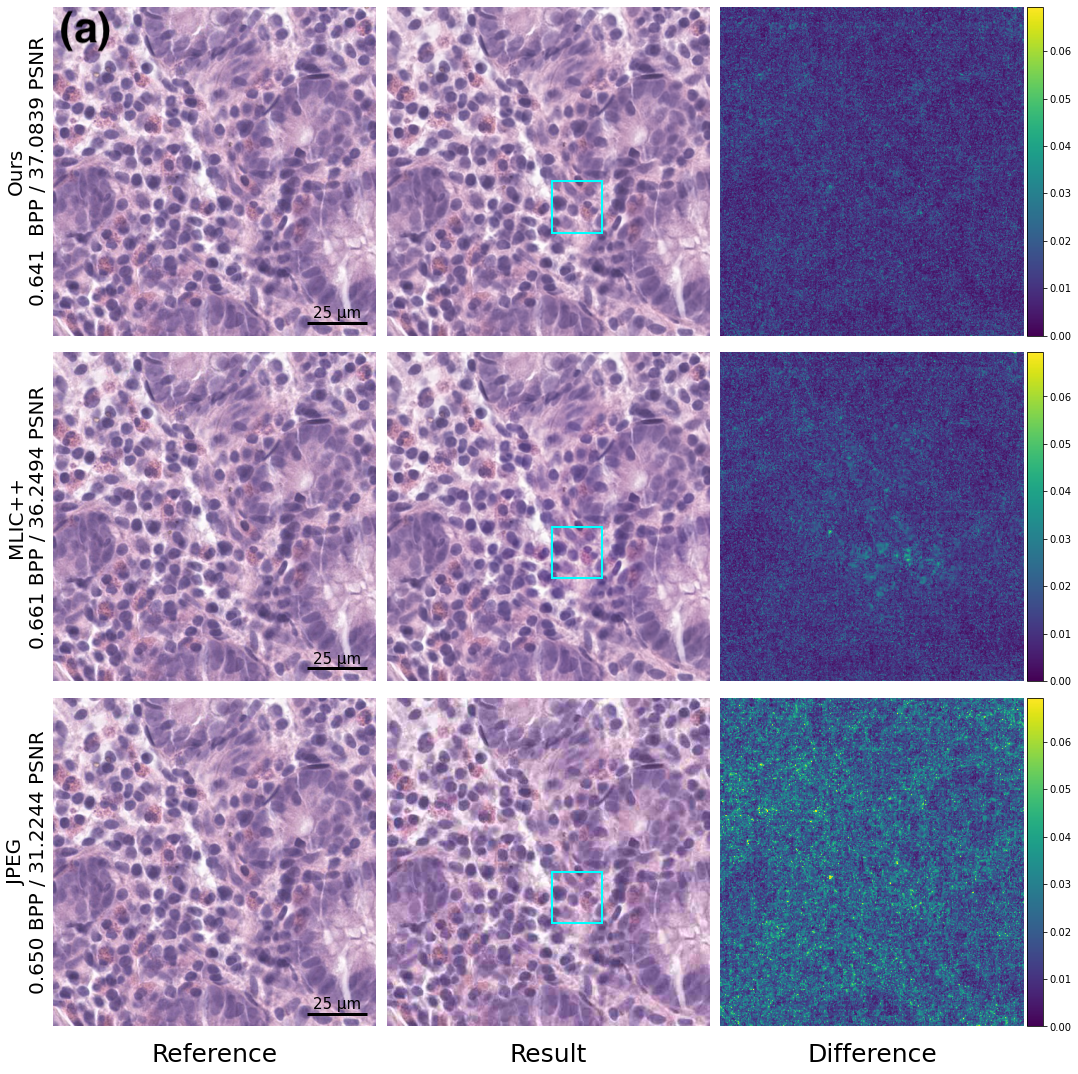}}
        \centerline{\includegraphics[width=\columnwidth ]{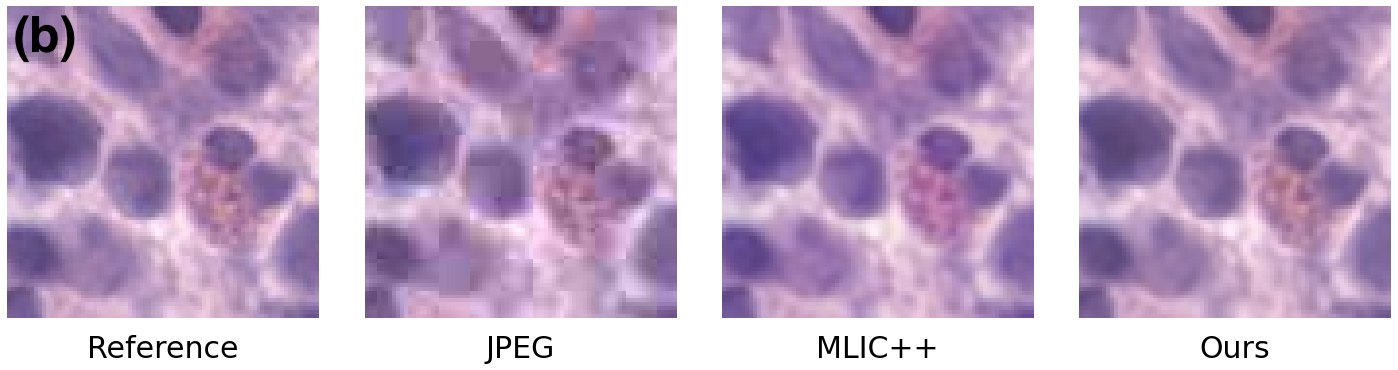}}
\vspace{-5pt}
\caption{Comparison of reconstruction results among CLERIC, MLIC++, and JPEG against the reference image. 
(a) The left column shows the original reference images, the middle column displays the reconstructed outputs, and the right column presents pixel-wise difference maps, where brighter regions indicate higher reconstruction errors. Each row corresponds to a different method: CLERIC (top), MLIC++ (middle), and JPEG (bottom). 
(b) Zoomed-in views of local regions highlighted by rectangular boxes in the second column of (a), illustrating detailed differences in reconstruction quality among the models.}
\vspace{-15pt}
        \label{fig:Pixel Difference}
\end{figure}
Figure~\ref{fig:Pixel Difference} (a) presents the reconstruction results of CLERIC, MLIC++ and JPEG when their compression rates were nearly identical, compared to the reference patch image (using a randomly selected test patch). 
This comparison highlights the ability of the proposed method to achieve more accurate reconstructions with lower pixel differences compared to the baseline, demonstrating its superior preservation of structural details.
In the magnified region shown in Figure~\ref{fig:Pixel Difference} (b), the proposed method demonstrates more accurate reconstruction of fine-grained tissue structures, further emphasizing its effectiveness in preserving subtle visual details.

\subsection{Ablation Study}

\begin{figure}
        \captionsetup{labelfont=bf}
        \centerline{\includegraphics[width=\columnwidth ]{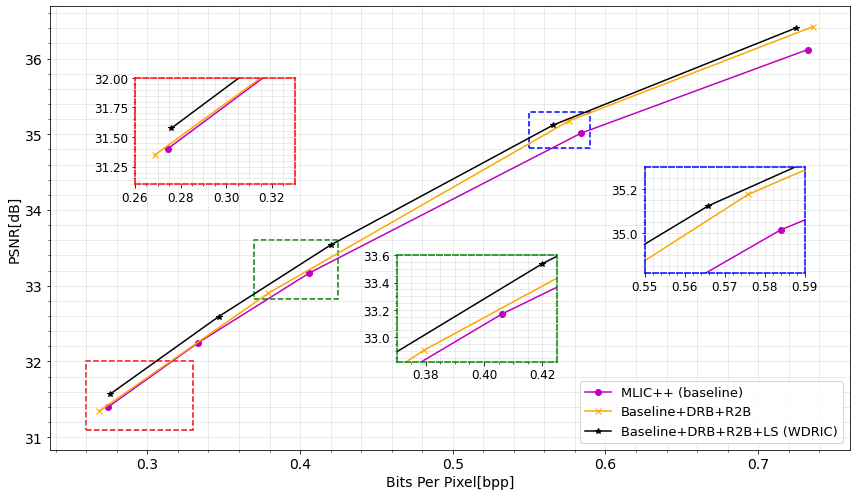}}
\vspace{-5pt}
        \caption{Ablation study results demonstrating the contribution of individual components to rate-distortion (RD) performance. The purple line represents the baseline model (MLIC++), the yellow line shows the effect of integrating advanced residual blocks (DRB $\&$ R2B), and the black line represents the final CLERIC model after further incorporating the lifting scheme (LS) module. The dotted squares highlight magnified regions to emphasize the performance gaps between configurations.}
\vspace{-20pt}
        \label{fig:ablation_ls}
\end{figure}

We conducted an ablation study to assess the contribution of each proposed module. Figure~\ref{fig:ablation_ls} and Table~\ref{tab:bd-rate_ablation_ls} illustrate the progressive improvement in the rate-distortion (RD) curve and BD-rate as individual modules are integrated, transitioning from the baseline model (MLIC++) to our proposed model (CLERIC). The results confirm that each module in both the encoder and decoder plays a significant role in enhancing RD performance.

As shown in Figure~\ref{fig:ablation_ls}, the lifting scheme module primarily improves performance at low $\lambda$ values (low bit-per-pixel, BPP) by enhancing compression efficiency, while the advanced residual blocks provide greater performance gains at high $\lambda$ values (high BPP) by refining reconstruction quality.

\begin{figure}
        \captionsetup{labelfont=bf}
        \centerline{\includegraphics[width=\columnwidth]{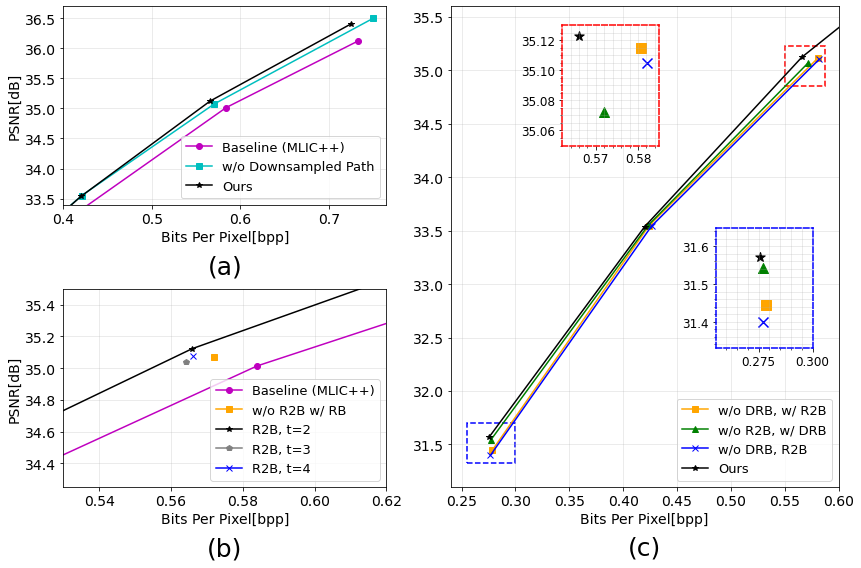}}
\vspace{-5pt}
        \caption{Ablation study results analyzing the impact of different architectural components on rate-distortion (RD) performance. (a) Effect of incorporating the downsampled patch image in the lifting scheme. (b) RD performance variations with different recursion iteration counts \( t \) in the Recurrent Residual Block (R2B). (c) Impact of removing Deformable Residual Blocks (DRB) and Recurrent Residual Blocks (R2B) individually or jointly.}
\vspace{-15pt}
        \label{fig:ablation_resblocks}
\end{figure}

Figure~\ref{fig:ablation_resblocks} (a) analyzes the effect of incorporating the downsampled patch image in the lifting scheme module, as illustrated in Figure~\ref{fig:CLERIC_framework}. The results indicate that leveraging both spatial and frequency domain information significantly improves RD performance, demonstrating the effectiveness of this hybrid representation.

Figure~\ref{fig:ablation_resblocks} (b) presents the RD performance variations with different recursion iteration counts $t$ in the Recurrent Residual Block (R2B). The results reveal that increasing $t$ improves performance up to $t = 2$, but beyond this point, the improvement becomes negligible or even slightly degrades. This suggests that setting $t = 2$ is optimal for balancing computational efficiency and RD performance.

Figure~\ref{fig:ablation_resblocks} (c) evaluates the RD impact of removing the Deformable Residual Block (DRB) and Recurrent Residual Block (R2B) individually or in combination. The results show that both DRB and R2B significantly contribute to performance enhancement, confirming that the deformable and recursive structures effectively improve feature extraction and compression quality.

In addition to RD performance, we further analyze the computational characteristics of each module, as presented in Table~\ref{tab:module_comparison}. This table summarizes the encoding/decoding latency, GFLOPs, and parameter counts for various module configurations.
While the integration of additional modules slightly increases encoding and decoding time, the results clearly show that these additions lead to significant improvements in RD performance.

\begin{table}
        \centering
        \caption{BD-Rate (\%) comparison in the ablation study, evaluated against VTM 17.0 Intra. Each module is progressively added to the baseline (MLIC++), demonstrating the performance improvements of Deformable Residual Blocks (DRB), Recurrent Residual Blocks (R2B), and the Lifting Scheme (LS) in the final CLERIC model.}
        \vspace{5pt}
        \label{tab:bd-rate_ablation_ls}
        \begin{tabular}{l r}
            \toprule
            \textbf{Method} & \textbf{BD-Rate (\%)} \\
            \midrule
            Baseline (MLIC++) & -19.78\% \\
            Baseline + DRB + R2B & -21.80\% \\
            Baseline + DRB + R2B + LS (CLERIC) & \textbf{-23.09\%} \\
            \bottomrule
        \end{tabular}
        \vspace{-10pt}
\end{table}

\begin{table}
        \centering
        \caption{Ablation studies on the effect of different module configurations on encoding/decoding latency, computational cost (GFLOPs), and network parameters. The computational cost and network parameters were measured during training phase, whereas the encoding and decoding latency was evaluated on in-house dataset in Figure~\ref{fig:RD_Curves}. The GFLOPs and Params were computed using calflops library\cite{calflops}.}
        \label{tab:module_comparison}
        \resizebox{\columnwidth}{!}{
        \begin{tabular}{c c c c c c c}
            \toprule
            \textbf{LS} & \textbf{DRB} & \textbf{R2B} & \textbf{Enc (s)} & \textbf{Dec (s)} & \textbf{GFLOPs} & \textbf{Params (M)} \\
            \midrule
            X & X & X & 0.1534 & 0.2147 & 75.44 & 83.5 M \\
            \checkmark & \checkmark & X & 0.1716 & 0.2437 & 64.06 & 85.87 M \\
            \checkmark & X & \checkmark & 0.1730 & 0.2406 & 91.85 & 85.34 M \\
            \checkmark & X & X & 0.1686 & 0.2367 & 81.18 & 85.34 M \\
            X & \checkmark & \checkmark & 0.1603 & 0.2270 & 88.71 & 83.94 M \\
            \checkmark & \checkmark & \checkmark & \textbf{0.1746} & \textbf{0.2467} & \textbf{74.72} & \textbf{85.87 M} \\
            \bottomrule
        \end{tabular}}
        
        \vspace{-15pt}
    \end{table}

\section{Discussion and Conclusion}
\label{sec:discussion_conclusion}

In this study, we proposed a novel deep learning-based image compression framework, CLERIC, designed specifically for whole slide images (WSIs) in digital pathology. Our model integrates a learned lifting scheme, deformable residual blocks (DRB), and recurrent residual blocks (R2B) to enhance compression efficiency while preserving fine-grained pathological details. By leveraging spatial-frequency decomposition and adaptive feature extraction mechanisms, CLERIC outperforms existing state-of-the-art (SOTA) learned image compression (LIC) models in rate-distortion (RD) performance. The model is particularly well-suited for the storage, transmission, and visualization of large-scale pathology images in clinical and research settings.

Our experimental results demonstrate that each proposed module contributes to performance improvement. The lifting scheme improves compression at low bit-per-pixel (BPP) values by efficiently capturing structured information, whereas the deformable and recurrent residual blocks enhance reconstruction accuracy at high BPPs by refining feature representations. The ablation study further confirms that combining spatial and frequency domain information, adaptive receptive fields, and iterative feature refinement significantly boosts RD efficiency. The final CLERIC model achieves a substantial BD-Rate gain over the baseline MLIC++, demonstrating its effectiveness for pathology image compression.

One key limitation of our experiment is that the ground-truth images used for training and evaluation were already subjected to JPEG compression by the scanners. This means that the reference images were not truly raw WSIs but were instead lossy images. In practice, if our deep learning-based compression were directly applied to uncompressed raw WSI data, the reconstructed images could achieve even higher fidelity than those typically observed by pathologists. Given the same compression ratio, this approach could potentially provide pathologists with clearer images than what is currently available from standard scanner outputs.

The primary goal of CLERIC is to reduce storage requirements while enabling the long-term preservation of large-scale pathology datasets. As shown in Figure~\ref{fig:RD_Curves}, the BPP gap between our model and JPEG is substantial, particularly at higher PSNR values. This indicates that CLERIC achieves superior compression while maintaining high reconstruction quality.

Inference speed is another crucial factor for real-time usability. When a user wishes to view a specific region of interest (ROI) on a monitor, only the patches corresponding to the selected resolution level need to be decompressed, as discussed in Section~\ref{sec:Methods}. For instance, if the monitor resolution is $1024\times2048$, only eight $512\times512$ patches need to be reconstructed, which takes approximately 1.97 seconds (calculated in Table~\ref{tab:module_comparison}). While this latency may cause a noticeable delay during zooming and navigation, the restoration time can be significantly reduced with more powerful GPU resources. Additionally, optimizing the decoding pipeline with algorithmic enhancements could further improve real-time responsiveness.

Although our results indicate strong compression performance, further qualitative evaluation from pathologists is necessary. Specifically, for disease diagnosis, it is critical to assess whether compression artifacts introduced by CLERIC negatively affect diagnostic accuracy in terms of sensitivity and specificity. Future studies should involve systematic assessments of diagnostic consistency between compressed and uncompressed WSIs across different pathological conditions.

Furthermore, beyond direct visualization, it is important to investigate whether the compressed WSIs impact the performance of deep learning models for disease classification and object detection. Many pathology AI models rely on patch-based analysis due to the enormous size of WSIs. Evaluating how CLERIC-compressed images influence downstream AI tasks will provide insights into the feasibility of integrating LIC methods into AI-assisted pathology workflows.

Another interesting research direction is the potential for directly utilizing compressed bitstreams in deep learning models for disease monitoring. Since WSIs are extremely large, conventional deep learning methods divide them into smaller patches for analysis, often treating each patch separately. However, the bitstreams generated by CLERIC are compact yet information-dense, containing most of the essential image details in a highly compressed form. If these bitstreams could be directly fed into deep learning networks, the models might learn to analyze pathology images more holistically, moving beyond localized patch-wise processing to a more global and context-aware representation of disease patterns.

In conclusion, our study demonstrates that CLERIC is a practical and effective deep learning-based compression framework for WSIs. By leveraging wavelet-based decomposition, adaptive feature extraction, and iterative refinement, the model achieves superior RD performance compared to existing methods. Its ability to significantly reduce storage requirements while maintaining diagnostic integrity makes it a promising solution for digital pathology applications, enabling efficient data management, long-term storage, and seamless integration into AI-driven pathology workflows.

\bibliographystyle{ieeetr}
\bibliography{reference}

\end{document}